\documentclass[conference]{IEEEtran}
\IEEEoverridecommandlockouts
\usepackage{cite}
\usepackage{amsmath,amssymb,amsfonts}

\usepackage{graphicx}
\usepackage{textcomp}
\usepackage{algorithm}
\usepackage{algpseudocode}
\usepackage{svg}
\usepackage{multirow}
\usepackage{hyperref}
\usepackage{amsmath}
\usepackage{blindtext}
\usepackage{xcolor}
\usepackage{float}
\usepackage{subcaption}

\def\BibTeX{{\rm B\kern-.05em{\sc i\kern-.025em b}\kern-.08em
    T\kern-.1667em\lower.7ex\hbox{E}\kern-.125emX}}
\begin{document}

\title{Cross Pseudo Supervision Framework for Sparsely Labelled Geospatial Images 
}
  
\author{\IEEEauthorblockN{Yash Dixit }
\IEEEauthorblockA{\textit{IUDX Programme Unit} \\
\textit{Indian Institute of Science}\\
Bengaluru \\
\href{https://orcid.org/0009-0002-5347-5715}{yashcdixit1998@gmail.com}}
\and
\IEEEauthorblockN{ Naman Srivastava}
\IEEEauthorblockA{\textit{IUDX Programme Unit} \\
\textit{Indian Institute of Science}\\
Bengaluru \\
\href{https://orcid.org/0000-0001-7521-6525}{srinaman2@gmail.com}}
\and
\IEEEauthorblockN{Joel D Joy }
\IEEEauthorblockA{\textit{IUDX Programme Unit} \\
\textit{Indian Institute of Science}\\
Bengaluru \\
\href{https://orcid.org/0009-0001-6179-9835}{djoeldjoy@gmail.com}}
\and
  \IEEEauthorblockN{Rohan Olikara}
 \IEEEauthorblockA{\textit{IUDX Programme Unit} \\
 \textit{ Indian Institute of Science}\\
 Bengaluru \\
 \href{mailto:rohan.olikara@gmail.com}{rohan.olikara@gmail.com}}
\and
 \hspace{1.15in} \IEEEauthorblockN{Swarup E}
\IEEEauthorblockA{\hspace{1.15in}\textit{IUDX Programme Unit} \\
\hspace{1.15in} \textit{Indian Institute of Science}\\
\hspace{1.15in} Bengaluru \\
\hspace{1.15in}  \href{mailto:swarup.e1998@gmail.com}{swarup.e1998@gmail.com}}
\and
\IEEEauthorblockN{Rakshit Ramesh}
\IEEEauthorblockA{\textit{IUDX Programme Unit} \\
\textit{Indian Institute of Science}\\
Bengaluru \\
\href{mailto:rakshit.ramesh@datakaveri.org }{rakshit.ramesh@datakaveri.org }}
}

\maketitle

\begin{abstract}
Land Use Land Cover (LULC) mapping is a vital tool for urban and resource planning, playing a key role in the development of innovative and sustainable cities. This study introduces a semi-supervised segmentation model for LULC prediction using high-resolution satellite images with a vast diversity of data distributions in different areas of India. Our approach ensures a robust generalization across different types of buildings, roads, trees, and water bodies within these distinct areas. We propose a modified  Cross Pseudo Supervision framework to train image segmentation models on sparsely labelled data. The proposed framework addresses the limitations of the famous 'Cross Pseudo Supervision' technique for semi-supervised learning, specifically tackling the challenges of training segmentation models on noisy satellite image data with sparse and inaccurate labels. This comprehensive approach significantly enhances the accuracy and utility of LULC mapping, providing valuable insights for urban and resource planning applications. 
\end{abstract}


% Redefine section numbering to use numbers
\renewcommand\thesection{\arabic{section}}
\renewcommand\thesubsection{\thesection.\arabic{subsection}}
\renewcommand\thesubsubsection{\thesubsection.\arabic{subsubsection}}

% Define command for subsubsubsection
\newcommand{\subsubsubsection}[1]{\paragraph{#1}}

\section{Introduction}

Satellite imagery plays a crucial role in sectors such as defence, resource management, environmental analysis, and urban planning. These images, varying in resolution and covering extensive areas, are essential for understanding and managing Earth's surface. Manual annotation for specific tasks requires significant human effort and time. To address this, researchers increasingly use deep learning techniques like object detection and image segmentation, offering automated and efficient geospatial data analysis. One important application is Land Use Land Cover (LULC) mapping, which enables governments and organizations to monitor resources, assess environmental changes, and formulate informed policies. Leveraging deep learning algorithms, LULC mapping can be performed more accurately and efficiently, aiding in sustainable development and resource management.

% \textbf{Problem Statement:}
Currently, land use class labels for urban planning, such as buildings, roads, and water bodies, rely on limited sources like manually annotated data from Java Open Street Maps \cite{OpenStreetMap} or model-generated predictions from Microsoft Building Footprints and Google Buildings. These sources are often sparsely labelled or focused on a single land use type, typically buildings. However, to foster intelligent and sustainable urban development, it is crucial to have a framework that enables the rapid and efficient generation of Land Use and Land Cover (LULC) maps, ensuring accurate capture of our dynamic and ever-evolving landscapes.

There have been several attempts to implement supervised and semi-supervised deep learning techniques for LULC prediction, addressing various aspects such as class imbalance and LULC for coastal areas and farmlands. \cite{chen2022semi,lu2022simple,cenggoro2017classification,xu2021semantic,sertel2022land,wang2021mask}. To create a robust LULC mapping model, it is crucial to account for variations in satellite images caused by factors like the time of day and date of the year. Additionally, the appearance of buildings and vegetation varies significantly across different regions in India due to diverse climatic, environmental, and social factors. Therefore, we need a model capable of accommodating these variations by training on a diverse dataset of satellite images from various regions and times of the year. A generalized model should perform well across geographical terrains such as coasts, plateaus, farmlands, river basins, and densely populated metropolitan cities.

Addressing these challenges, this study works on the following research question: \\
\textbf{RQ:} Creating a generalized framework Semantic Segmentation model to enhance the performance of Deep Learning models to predict various land use classes present in a satellite image. \\

\section{Background and Related Works}

\subsection{Semantic Segmentation}

\subsubsection{U-Net}
The U-Net architecture, primarily designed for biomedical image segmentation\cite{ronneberger2015u}, has become highly influential in land use classification tasks \cite{jiwani2021semantic,robinson2022fast,sirko2021continental}.
% At its core, U-Net, with a U-shaped architecture, consists of two paths: a contracting path(encoder) and an expansive path(decoder).
\subsubsection{DeepLab v3+}
    DeepLabV3+, an advanced deep learning model developed by Google DeepMind \cite{chen2018encoder}, has significantly enhanced land use classification tasks. This model utilizes a convolutional neural network (CNN) architecture featuring key components such as Atrous Spatial Pyramid Pooling (ASPP) and an encoder-decoder structure. The ASPP module is designed to capture multi-scale context by applying atrous (dilated) convolutions at multiple scales, which helps the model understand spatial hierarchies in images. The encoder-decoder structure further refines this by ensuring detailed spatial information is preserved, making DeepLabV3+ particularly effective for classifying various land use types from satellite or aerial imagery.\cite{jiwani2021semantic,fan2022land}.  
\subsection{Semi-Supervised Semantic Segmentation }
\subsubsection{Cross Pseudo Supervision (CPS)}
CPS introduces a consistency regularization approach with two segmentation networks initialized differently. It enforces consistency by using one network's pseudo-one-hot label map to supervise the other, and vice versa, promoting uniform predictions \cite{chen2021semi}.

\begin{figure*}[h]
    \centering
    \includegraphics[width=1\textwidth ]{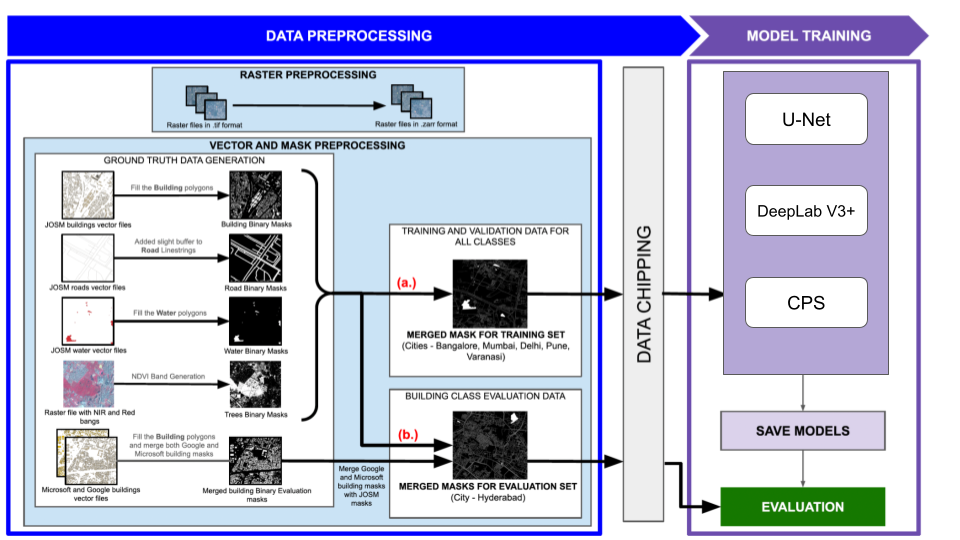}
    \caption{Workflow for LULC Segmentation. \textcolor{red}{\textbf{(a.)}}:  The Merged Training masks are created by combining the binary class masks of the concerned cities which have been generated by using the JOSM vector files. Apart from these classes, the remaining areas are classified as the "Other" class. During training, we focus on a subset of data that is densely populated with classes and extract chips from it.
    \textcolor{red}{\textbf{(b.)}}: The Merged Evaluation masks are created by utilizing the output binary masks of roads, water, trees, combined data of buildings (JOSM, Microsoft, and Google), and treating the remaining areas as the "Other" class. During the model evaluation, data is chipped without imposing class-heavy constraints.}
    \label{fig:pipeline}
\end{figure*}

\section{Methodology}
The proposed model architecture unfolds through several essential stages. Initially, data preparation involves the creation of raster masks from vector data associated with each satellite image. Figure \ref{fig:pipeline} depicts the deployed machine learning operations pipeline by illustrating the sequence of these stages and how they are connected to achieve the desired outcome.

\subsection{Data Sources}
\subsubsection{Raster Data}
In this study, we utilized six distinct high-resolution (around $1.134m ^ {2}$/px) NRGB (Near-Infrared, Red, Green, and Blue Bands) satellite image scenes obtained from the Indian Space Research Organisation's (ISRO) CartoSAT-3 satellite multispectral instrument.
\subsubsection{Vector Data}
 The Vector Data is acquired for training and evaluation sets differently.\\
- \textbf{Training : } We acquired vector files for various land uses, which are enclosed within the extent of the geospatial coordinates of raster files for the cities of Bangalore, Mumbai, Pune, Varanasi and Delhi. \\
- \textbf{Evaluation : }
 For our evaluation set, we included data from Google Buildings and Microsoft Footprints for buildings and the JOSM vector for the remaining classes. The evaluation data consists of data from a small region in the city of Hyderabad.

\subsection{Data Preparation}
\subsubsection{Class-wise Binary Mask Generation for Segmentation}
To generate a binary mask for each class, we used GeoJSON files containing vector geometries such as Linestrings (for roads), Polygons, and multi-polygons (for buildings and water bodies). Sparse labels for all classes were acquired from JOSM. For roads, we applied a 3-pixel buffer around the Linestrings to ensure the masks covered visible road areas in the satellite images. For Polygons and Multipolygons representing buildings and water bodies, we filled the entire polygon with the value 1. The processes of dilation and filling are detailed in Figure \ref{fig:pipeline}. The vegetation mask was generated using the Normalized Difference Vegetation Index (NDVI), with values above -0.1 designated as 1 and those below as 0.

\subsubsection{Data Chipping} 
In this step, we create chips, also known as patches, from the Zarr file by breaking down the large raster into smaller chunked images of consistent size (4 x 256 x 256 each). These chips are generated for both the Cartosat images and their corresponding masks. To ensure high-quality data for training, we apply a unique filter during chip generation. We select only the Cartosat and its corresponding mask chip in the training dataset whenever the Cartosat chip has less than 50\% of the area marked as NaN. For evaluation, the entire Area of Interest is considered.

\subsection{Model Training}
Our comparative study evaluates both supervised and semi-supervised techniques. For supervised learning, we use Unet and DeepLabV3+ as baseline model architectures. For semi-supervised methods, we incorporate Cross Pseudo Supervision.
\subsubsection{Vanilla UNet}
The UNet model architecture from the Segmentation Models Pytorch library serves as a baseline. It is trained using a weighted pixel-wise cross-entropy loss to address class imbalance, with class weights inversely proportional to the abundance of each class in the dataset \cite{Iakubovskii:2019}.

\subsubsection{Vanilla DeepLabv3+}
We implement DeepLabV3+ with the EfficientNet backbone and Tversky Loss for another supervised learning model.

\begin{figure*}[!ht]
    \centering
    \includegraphics[width=1\textwidth]{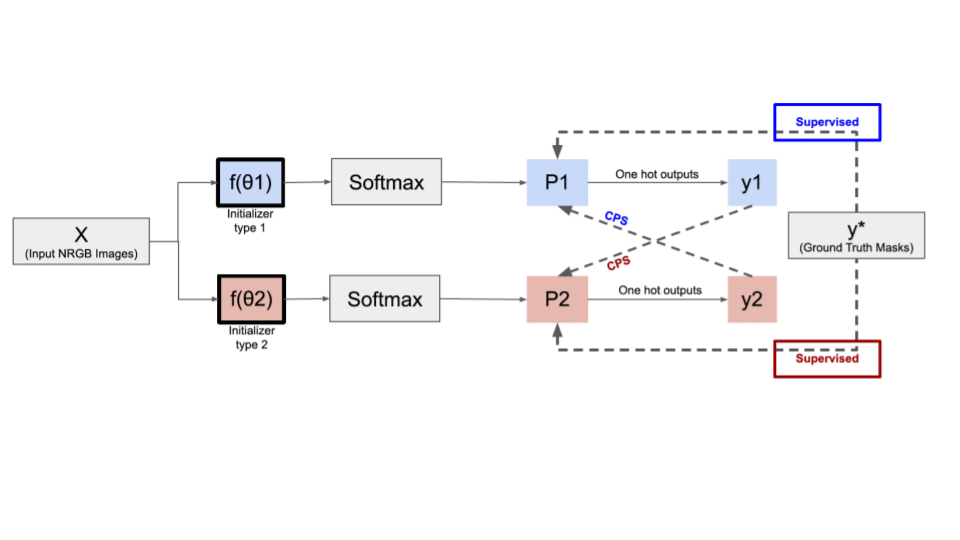}
    \caption{ Cross Pseudo Supervision Architecture. $f(\theta1)$ and $f(\theta2)$ are the 2 distinct Deeplabv3+ models which are being trained. P1 and P2 are the logits that have been generated from the respective models and y1 and y2 are their corresponding one-hot encoded values. $y^*$ are the ground truth models}
    \label{fig:arch}
\end{figure*}

\subsubsection{Cross Pseudo Supervision (CPS)}
In the cross-pseudo supervision method, we leverage two DeepLabV3+ models, each with EfficientNet backbones and distinct initializations, as depicted in Figure \ref{fig:arch} with the models represented by \( f(\theta1) \) and \( f(\theta2) \). The training data is characterized by sparse labelling, which permits the computation of both supervised and unsupervised losses from the same batch of labelled data. This methodology diverges from the approach described by Xiaokang Chen et al. (2021) \cite{chen2021semi}, as in this study the model is trained over sparsely labelled data in contrast to the use of a separate labelled and unlabelled dataset, as demonstrated by \cite{chen2021semi}.
The proposed approach begins by calculating the supervised loss through the weighted Hausdorff Erosion loss \cite{HausdorffER} for each ground truth versus predicted class across five classes, as detailed in Equation \ref{eq:Haursd}. The Hausdorff loss per pixel per class denoted as \( \text{\textit{l}}_{hd} \), is determined for both models and then aggregated. This aggregated Hausdorff Loss is designated as \( L^{HF}_{Sup} \). Subsequently, the Weighted Cross Entropy Loss, symbolized as \( l_{\text{ce}} \), is calculated using the ground truth and predicted masks for both models, as defined in Equation \ref{eq:wce_sup}, and then combined. The cumulative Weighted Cross Entropy loss is represented by \( L^{WCE}_{Sup} \).
Both Equations \ref{eq:Haursd} and \ref{eq:wce_sup} are multiplied by the reciprocal of a constant \( K \), which is defined as the product of \( |D_{\text{SL}}| \) (the number of sparsely labelled image samples in the training set), \( W \) (the width), and \( H \) (the height of the images). The total supervised loss, \( L_{Supervised} \), is computed by integrating the aggregated Hausdorff and Weighted Cross Entropy losses, with the cross-entropy loss being weighted less by a factor of 0.5, as illustrated in Equation \ref{eq:sup}.

\begin{equation}
    \begin{aligned}
        K &= {|D_{\text{SL}}|} \times {W\times H}
    \end{aligned}
    \label{eq:constant}
\end{equation}

\begin{equation}
    \begin{aligned}
        L^{HF}_{Sup} &= \frac{1}{K} \times \sum_{X\in D_{\text{SL}}} \sum_{w=1}^{5} \sum_{i=0}^{W\times H}  \alpha_w \times \bigl( l_{\text{hd}}(P_{1iw}, y^{*}_{iw})  \\
        & \hspace{4cm} + l_{\text{hd}}(P_{2iw}, y^{*}_{iw}) \bigr)
    \end{aligned}
    \label{eq:Haursd}
\end{equation}

\begin{equation}
    \begin{aligned}
        L^{WCE}_{Sup} &= \frac{1}{K} \times \sum_{X\in D_{\text{SL}}} \sum_{w=1}^{5} \sum_{i=0}^{W\times H}  \alpha_w \times  \bigl( l_{\text{ce}}(P_{1iw}, y^{*}_{iw})  \\ 
        & \hspace{4cm} + l_{\text{ce}}(P_{2iw}, y^{*}_{iw})  \bigr)
    \end{aligned}
    \label{eq:wce_sup}
\end{equation}

\begin{equation}
    L_{Supervised} = L^{HF}_{Sup} + 0.5* L^{WCE}_{Sup}
    \label{eq:sup}
\end{equation}

We exclusively employ weighted cross-entropy loss for cross pseudo supervision that is calculated between the softmax outputs from model2 and the one hot outputs generated from softmax outputs of model1 and vice versa, as described in Equation \ref{eq:wce_cps}. Then the losses for both models are summed together. This total Cross Pseudo supervision loss is denoted as $L^{WCE}_{CPS}$

\begin{equation}
    \begin{aligned}
        L^{WCE}_{CPS} &= \frac{1}{K} \times \sum_{X\in D_{\text{SL}}} \sum_{w=1}^{5} \sum_{i=0}^{W\times H}  \alpha_w \times \bigl( l_{\text{ce}}(P_{1iw}, y_{2iw}) \\ 
        & \hspace{4cm} +  l_{\text{ce}}(P_{2iw}, y_{1iw}) \bigr)
    \end{aligned}
    \label{eq:wce_cps}
\end{equation}

$P_{1iw}$ and $P_{2iw}$ are the logit outputs from the two different models, $y^{*}_{iw}$ are the ground truth masks, and $y_{1iw}$ and $y_{2iw}$ are the one hot vectors computed from the corresponding $P_{1iw}$ and $P_{2iw}$.

Following all of the steps, we determine the total loss by adding $L_{\text{Supervised}}$ and $L^{\text{WCE}}_{\text{CPS}}$. To account for the evolving significance of the unsupervised loss component within the total loss, we apply a sigmoid ramp-up to adjust the lambda trade-off. This is achieved by gradually increasing the weightage of $L^{\text{WCE}}_{\text{CPS}}$ in Equation \ref{eq:totalLoss} by incrementing $\lambda$ from 0 to 0.1 across the specified number of epochs as specified in Equation \ref{eq:rampup} and detailed by Laine et. al. (2016) \cite{laine2016temporal}.

\begin{equation}
    L_{Total} = L_{Supervised} + \lambda*L^{WCE}_{CPS}
    \label{eq:totalLoss}
\end{equation}

\begin{equation}
    \text{rampup}(r, t, T)=
\begin{cases} 
0 & \text{if } t = 0, \\
0.1*\exp\left(-5.0 \left(1.0 - \frac{t}{T}\right)^2\right) & \text{if } \text{r$\ge$t$\ge$1}, \\
0.1 & \text{otherwise}.
\end{cases}
\label{eq:rampup}
\end{equation}

The trade-off parameter $\lambda$ is ramped up using the Equation \ref{eq:rampup} where \textit{T} is the Total number of epochs, \textit{t} is the current epoch, and \textit{r} is the ramp up length. \\

\begin{table*}[ht]
    \centering
    \caption{Recall Score For Supervised and Semi-Supervised Model}
    \begin{tabular}{|p{2.1cm}|c|c|c|c|c|}
    \hline
        \textbf{Model/Framework} & \textbf{Threshold} & \textbf{Trees} & \textbf{Buildings} & \textbf{Water} & \textbf{Roads} \\
        \hline
        \hline
    
         \multirow{2}{*}{CPS} & 0.4 & \textbf{82.7912} & \textbf{75.5499} & \textbf{76.9064} & \textbf{66.6928} \\
         \cline{2-6}
          & 0.5 & 78.5390 & 65.1801 & 66.0376 & 58.4839 \\
         \hline
         \multirow{2}{*}{Unet} & 0.4 & 12.5973 & 45.6297 & 62.059 & 0.1026 \\
         \cline{2-6}
          & 0.5 & 12.368 & 45.0844 & 61.785 & 0.0968 \\
         \hline
         \multirow{2}{*}{Deeplabv3+} & 0.4 & 18.0161 & 47.0938 & 60.9702 & 3.2851 \\
         \cline{2-6}
          & 0.5 & 16.6227 & 53.2109 & 70.0857 & 0.0410 \\
         \hline
    \end{tabular}
    
    \label{tab:recall-scores}
\end{table*}

\subsection{Post Processing} We applied two techniques to post-process the outputs obtained from the two models:\\
- \textbf{Prediction Ensembling}:
 The final prediction for each chip is an average of the softmax probabilities from both models.\\
- \textbf{Prediction Merging}:
We merge output chips based on geospatial coordinates using max pooling to recreate the full image. This method leverages overlapping chips to ensure broader context, enhancing evaluation scores and predictions.

\section{Results and Discussion}

We present the results from various experiments using both supervised and semi-supervised approaches. Since DeepLabV3+ outperforms U-Net, we use it as the base model for evaluating CPS frameworks within the semi-supervised approach. The recall scores for these models, calculated on a subsection of the Hyderabad Cartosat image covering $24.6km^{2}$, are shown in Table \ref{tab:recall-scores}.

Due to sparse ground truth labels, evaluating our model's performance using standard metrics like Accuracy or Dice score is challenging as these metrics often misclassify predictions against the sparse labels. Instead, we use Recall, which emphasizes True Positives and False Negatives, making it more suitable for this context. This approach ensures proper assessment by minimizing False Negatives and maximizing True Positives.

Table \ref{tab:recall-scores} shows that the Cross Pseudo Supervision (CPS) based semi-supervised approach significantly outperforms single supervised models like DeepLabV3+ and U-Net, with an average Recall score of 75.5 across all classes. This method effectively addresses issues caused by sparse and inaccurate labels in an ever-changing landscape. 
There is a scope for further improvement in the model's performance by addressing issues such as dynamic weighting strategies for loss function to address the class imbalance. We can also improve the model performance by applying cloud removal and atmospheric correction techniques to the geospatial images. This improves the input data quality, thereby affecting the overall model performance.  

% Based on Table \ref{tab:recall-scores}, our study concludes that Cross Pseudo Supervision provides a significant advantage over Supervised Learning methods for Land Use Land Cover classification.

\section{Conclusion}
This study assessed both supervised and semi-supervised techniques for Land Use Land Cover (LULC) Segmentation, employing DeepLabV3+ and U-Net as foundational models. The semi-supervised technique incorporating Cross Pseudo Supervision (CPS) exhibited enhanced efficacy relative to the supervised models. The CPS-based approach proficiently managed challenges stemming from sparse and imprecise labels, demonstrating significant adaptability to diverse and evolving landscapes. Prospective improvements include the implementation of dynamic weighting strategies within the loss function to mitigate class imbalance, alongside the adoption of cloud removal and atmospheric correction methodologies to elevate the quality of geospatial imagery, thereby further refining the model's performance.

\bibliographystyle{unsrt} % We choose the "plain" reference style
\bibliography{main} 

\end{document}